\documentclass{article} 
\usepackage{iclr2015,times}
\usepackage{hyperref}
\usepackage{url}
\usepackage{float}
\usepackage{layout}
\usepackage{amsmath}
\usepackage{amsfonts}
\usepackage{graphicx}
\usepackage{multirow}
\usepackage{makecell}
\usepackage[export]{adjustbox}

\newcommand{\comment}[1]{}

\title{Pixel-wise Deep Learning for Contour Detection}

\author{
Jyh-Jing Hwang \& Tyng-Luh Liu \\
Institute of Information Science\\
Academia Sinica\\
Taipei, Taiwan \\
\texttt{\{jyhjinghwang,liutyng\}@iis.sinica.edu.tw}
}

\iclrfinalcopy 

\begin{document}

\maketitle

\begin{abstract}
We address the problem of contour detection via per-pixel classifications of edge point. To facilitate the process, the proposed approach leverages with DenseNet, an efficient implementation of multiscale convolutional neural networks (CNNs), to extract an informative feature vector for each pixel and uses an SVM classifier to accomplish contour detection. In the experiment of contour detection, we look into the effectiveness of combining per-pixel features from different CNN layers and verify their performance on BSDS500.
\end{abstract}

%
\section{Introduction}
\label{sec:intro}
%

We consider deep nets to construct a per-pixel feature learner for contour detection. As the task is essentially a classification problem, we adopt deep convolutional neural networks (CNNs) to establish a discriminative approach. However, one subtle deviation from typical applications of CNNs should be emphasized. In our method, we intend to use the CNN architecture, e.g., AlexNet~\citep{alexnet}, to generate features for each image pixel, not just a single feature vector for the whole input image. Such a distinction would call for a different perspective of parameter fine-tuning so that a pre-trained per-image CNN on ImageNet \citep{Deng2009imagenet} can be adapted into a new model for per-pixel edge classifications. To investigate the property of the features from different convolutional layers, we carry out a number of experiments to evaluate their effectiveness in performing contour detection on the benchmark BSDS Segmentation dataset~\citep{bsds}.

%
\section{Per-Pixel CNN Features}
\label{sec:model}
%

Learning features by employing a deep architecture of neural net has been shown to be effective, but most of the existing techniques focus on yielding a feature vector for an input image (or image patch). Such a design may not be appropriate for vision applications that require investigating image characteristics in pixel level. For contour detection, the central task is to decide whether an underlying pixel is an edge point or not. Thus, it would be convenient that the deep network could yield per-pixel features. To this end, we extract per-pixel CNN features in AlexNet~\citep{alexnet} using DenseNet~\citep{densenet}, and pixel-wise concatenate them to feed into a support vector machine (SVM) classifier.

Our implementation uses DenseNet for CNN feature extraction owing to its efficiency, flexibility, and availability. DenseNet is an open source system that computes dense and multiscale features from the convolutional layers of a Caffe CNN based object classifier. The process of feature extraction proceeds as follows. Given an input image, DenseNet computes its multiscale versions and stitches them to a large plane. After processing the whole plane by CNNs, DenseNet would unstitch the descriptor planes and then obtain multiresolution CNN descriptors.

The dimensions of convolutional features are ratios of the image size, e.g., one-fourth for Conv1, and one-eighth for Conv2. We rescale feature maps of all the convolutional layers to the image size. That is, there is a feature vector in every pixel. The dimension of the resulting feature vector is $1376\times1$, which is concatenated by Conv1 ($96\times1$), Conv2 ($256\times1$), Conv3 ($384\times1$), Conv4 ($384\times1$), and Conv5 ($256\times1$).

For classification, we use the combined per-pixel feature vectors to learn a binary linear SVM. It is worth noting that, in our multiscale setting, we train the SVM based on only the original resolution. In test time, we classify test images using both the original and the double resolutions. We average the two resulting edge maps for the final output of contour detection.

%
\section{Experimental Results}
\label{sec:exp}
%

We test our method on the Berkeley Segmentation Dataset and Benchmark (BSDS500) \citep{bsds,gpb}. To better assess the effects of the features of different layers, we report their respective performance of contour detection. The BSDS500 dataset contains 200 training, 100 validation, and 200 testing images. Boundaries in each image are labeled by several workers and are averaged to form the ground truth. The accuracy of contour detection is evaluated by three measures: the best F-measure on the dataset for a fixed threshold (ODS), the aggregate F-measure on the dataset for the best threshold in each image (OIS), and the average precision (AP) on the full recall range \citep{gpb}. Prior to evaluation, we apply a standard non-maximal suppression technique to edge maps to obtain thinned edges \citep{canny}.

In Table~\ref{tbl:joint}, we see that features in Conv2 contribute the most, and then Conv3 and Conv4. These suggest that low- to mid-level features are most useful for contour detection, while the lowest- and higher-level features provide additional boost. Although features in Conv1 and Conv5 are less effective when employed alone, we achieve the best results by combining all five streams. It indicates that the local edge information in low-level features and the object contour information in higher-level features are
both necessary for achieving high performance in contour detection tasks.

\begin{table}[tH]
\caption{ Contour detection results of using CNN features from different layers.}
\label{tbl:joint}
\centering
\renewcommand{\arraystretch}{1.5}
\addtolength{\tabcolsep}{+8pt}
\small
\begin{tabular}{@{}l@{\hskip 0.3cm}l@{\hskip 0.3cm}l@{\hskip 0.3cm}l@{\hskip 0.3cm}l@{\hskip 0.3cm}l@{\hskip 0.3cm}l@{}}
{\normalsize \textbf{}} & Conv1 & Conv2 & Conv3 & Conv4 & Conv5 & Conv1-5 \\ \Xhline{3\arrayrulewidth}
ODS & .627 & .699 & .655 &.654 & .604 & .741 \\
OIS & .660 & .718 & .670 &.667 & .620 & .759 \\
AP  & .625 & .712 & .619 &.615 & .546 & .757 \\
\end{tabular}
\end{table}

\subsubsection*{Acknowledgments}
This work was supported in part by NSC 102-2221-E-001-021-MY3.

\bibliography{CSCNN}
\bibliographystyle{iclr2015}

\end{document}